\documentclass[conference]{IEEEtran}
\IEEEoverridecommandlockouts

\usepackage{cite}
\usepackage{amsmath,amssymb,amsfonts}
\usepackage[linesnumbered,ruled,vlined, algo2e]{algorithm2e}
\usepackage{algorithm, algpseudocode}
\usepackage{graphicx}
\usepackage{textcomp}
\usepackage{xcolor}
\usepackage[utf8]{inputenc}
\usepackage{kotex}
\usepackage{kotex-logo}

\def\BibTeX{{\rm B\kern-.05em{\sc i\kern-.025em b}\kern-.08em
    T\kern-.1667em\lower.7ex\hbox{E}\kern-.125emX}}

\begin{document}

\title{Personalized Federated Learning with Clustering: Non-IID Heart Rate Variability Data Application}

\newcommand*{\affaddr}[1]{#1}
\newcommand*{\affmark}[1][*]{\textsuperscript{#1}}
\newcommand*{\email}[1]{\texttt{#1}}

\author{Joo Hun Yoo\affmark[1], Ha Min Son\affmark[1], Hyejun Jeong\affmark[1],
        Eun-Hye Jang\affmark[2], Ah Young Kim\affmark[2], Han Young Yu\affmark[2], \\Hong Jin Jeon\affmark[3,*], 
        Tai-Myoung Chung\affmark[1,*]\\
        
\affaddr{\small \affmark[1] College of Computing and Informatics, Sungkyunkwan University}\\
\affaddr{\small \affmark[2] Bio-Medical IT Convergence Research Division, Electronics and Telecommunications Research Institute}\\
\affaddr{\small \affmark[3] Department of Psychiatry, Samsung Medical Center}\\
\email{\footnotesize \{andrewyoo, sonhamin, june.jeong\}@g.skku.edu}
\email{\footnotesize \{cleta4u, aykim, uhan0\}@etri.re.kr}\\
\email{\footnotesize \{jeonhj, tmchung\}@skku.edu}
}

\maketitle

\begin{abstract}

While machine learning techniques are being applied to various fields for their exceptional ability to find complex relations in large datasets, the strengthening of regulations on data ownership and privacy is causing increasing difficulty in its application to medical data. In light of this, Federated Learning has recently been proposed as a solution to train on private data without breach of confidentiality. This conservation of privacy is particularly appealing in the field of healthcare, where patient data is highly confidential.
However, many studies have shown that its assumption of Independent and Identically Distributed data is unrealistic for medical data. In this paper, we propose Personalized Federated Cluster Models, a hierarchical clustering-based FL process, to predict Major Depressive Disorder severity from Heart Rate Variability. By allowing clients to receive more personalized model, we address problems caused by non-IID data, showing an accuracy increase in severity prediction. This increase in performance may be sufficient to use Personalized Federated Cluster Models in many existing Federated Learning scenarios.

\end{abstract}

\begin{IEEEkeywords}
Federated Learning, Non-IID, clustering, personalized model, healthcare data, Major Depressive Disorder
\end{IEEEkeywords}

\section{Introduction} \label{sec:intr}

Machine Learning (ML) techniques are being applied to various research fields for their ability to learn from large amounts of data and extract meaning information \cite{chen2020deep}. This is advantageous, as the various statistical analysis methods used in the past, such as logistic regression analysis, have the limitation of inconsistent performance considering extensive number of variables. Deep learning (DL) techniques in particular are being used to find complex relations between both dependent and independent variables in large sets of data---a highly time consuming task for human experts \cite{tobore2019health}. However, the full transition to DL in the field of healthcare has some practical difficulties. 

While deep learning techniques are assumed to have a large, accessible, and centralized dataset, it is becoming difficult to apply DL to healthcare data due to restrictions of privacy laws. Many organizations have reported changes to their data privacy regulations~\cite{houser2018gdpr}, such as the tightening rules of EU's General Data Protection Regulation (GDPR), California Consumer Privacy Act (CCPA), and China's Cyber Security Law (CSL). In light of this, Federated Learning (FL) has been introduced to solve the limitation of centralized DL approaches. FL can comply with strict privacy regulations, as it does not rely on data collection. Owning to these advantages, FL has potential in its application to the healthcare field. However, FL is also known to have limitations due to its assumption of Independent and Identically Distributed (IID) data. In this research, we present a method to alleviate this assumption with an applications of FL in the medical field, particularly in the prediction of Major Depressive Disorder (MDD) severity.

MDD is a mental disorder characterized by at least two weeks of pervasive low mood, loss of interest, and low energy \cite{belmaker2008major}. Current evaluation of depression symptoms relies on scoring methods such as Hamilton Depression (HAM-D), Montgomery–Åsberg Depression Rating Scale (MADRS), and Beck's Depression Inventory (BDI), with more specific diagnosis through consultation with psychiatrists. Nevertheless, these diagnostic methods are significantly dependent on the mood of the subject and requires a considerable amount of time to conduct. As a solution, researchers have analyzed diagnosis of MDD using biomarkers such as neuroimage, Electroencephalogram (EEG), and Heart Rate Variability (HRV). We focus on HRV for its advantage of having a lower data collection cost, and the simplicity of its measurement method.

In this paper, we use FL with hierarchical clustering to create Personalized Federated Cluster Models (PFCM). Previous researches regarding mental healthcare analysis were conducted without consideration to data privacy; this creates difficulties applying DL methods to practical scenarios. Other studies that have addressed the concerns for data privacy overlooked the presentation of results in a realistic non-IID environment. In this research, we use FL as a method to leverage the strengths of DL while preserving data privacy. Clustering is used as a method to supplement the limitations of FL in non-IID environment---as is the case with healthcare dataset. Our main contributions are summarized as follows:

\begin{itemize}

\item We propose a privacy-preserving framework that leverages the strengths of deep learning technique in MDD diagnosis.
\item We improve HRV-based analytical performance of MDD severity by creating Personalized Federated Cluster Models.

\end{itemize}

\section{Background \& Related works} \label{sec:related}

\subsection{Federated Learning and Clustering Methods}

Federated Learning (FL) ~\cite{mcmahan2017communication, konevcny2016federated} is a distributed ML framework that allows training on data distributed across multiple clients without breach of confidentiality. One of the fundamental aspects of FL is privacy preservation, as there is no need for data collection. 
FL typically progresses with the repetitions of the following steps. A centralized server creates an initial global model and distributes it to each client. Each client trains its copy of the model with local data. Then, each client sends the local updates of the model to the server. The server aggregates the clients' updates, modifies the global model, and distributes the updated global model back to the clients. These steps are repeated until the model converges, or a stopping criterion is met~\cite{Li_2020}. By doing so, training is achieved exclusively through local model updates, without access to raw data, resulting in attack surface reduction~\cite{mcmahan2017communication}.

\begin{figure}[!ht]
\centering
\includegraphics[width=0.95\linewidth]{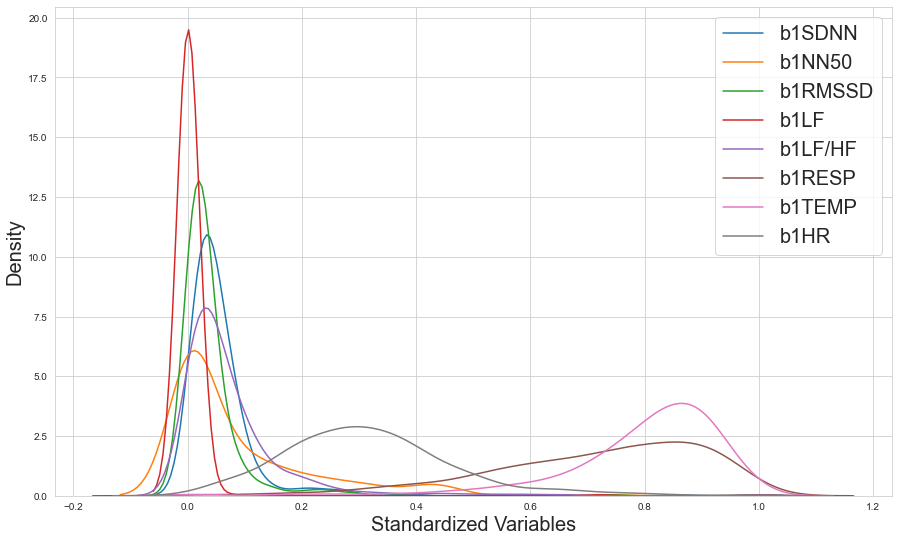}
\caption{Distributional differences between HRV dataset variables. Values have non-identical distribution even though they have undergone a scale standardization process. This specifically represents one of the concept shift, same label but different feature, where conditional distributions may vary.}
\label{fig:rand_var}
\end{figure}

While conventional FL algorithms, such as FedAvg~\cite{mcmahan2017communication}, show promising results, it relies heavily on an environment in which the data spread across clients are Independent and Identically Distributed (IID). Real-world healthcare data, however, are by nature, non-IID, heterogeneously distributed across the clients---an environment in which each random variable does not agree with the same probability distribution, and not all variables are mutually independent~\cite{Li_2020}. Figure \ref{fig:rand_var} shows the distribution of variables measured in baseline phase among the variables in HRV data used in this study. Not only for the differences in personal characteristics, but features extracted from the same HRV variables also have a non-identical distribution. In short, conventional FL algorithms cannot provide an optimal solution, as it is built on the assumption of IID data.

Many researchers have attempted to address this non-IID issue by blending different approaches in Computer Science to FL. References~\cite{sattler2020clustered, briggs2020federated} combined different clustering algorithms at many different steps in training. By grouping clients with similarities, different models for each cluster of clients are created, resulting in a reduced non-IIDness of data within a single cluster. Reference ~\cite{sattler2020clustered, briggs2020federated} resembles our work in terms of technical methods, as they leverage the hierarchical clustering algorithm to alleviate the ambiguity of pre-defining a number of clusters. However, the listed studies do not fully leverage the strength of clustering. During the testing phase, clients in the test set are allocated to all cluster models rather than distributed to their closest respective cluster. 

\subsection{HRV data for Major Depressive Disorder Diagnosis}

Heart Rate Variability (HRV) has been gaining attention in the field of depression analysis for its prominent correlation with depressive symptoms. Most previous researches chose the Support Vector Machine with feature selection (SVM-RFE) algorithm as the key ML algorithm. Table \ref{tab1} summarizes some significant related works. Reference \cite{kim2017diagnosis} applied the SVM-RFE algorithm to HRV, serum proteomics, and other multi-modal data. Reference \cite{byun2019detection} proposed an MDD diagnosis system using SVM-RFE only on HRV.
Reference \cite{kobayashi2019} proposed a single-lead ECG system based on HRV reactivity, ECG signal, and mental tasks analysis, to classify MDD subjects from non-MDD subjects. 
Most of these studies report promising results. 
Their approaches, however, are limited as classifying MDD into only two classes do not fully reflect its non-binary nature. As such, patients with mild or borderline depression scores may be inaccurately placed into different classes.
In addition, they do not address the problem of privacy regulations. Like most ML algorithms, SVMs rely on the premise of a centralized and accessible set of data. In practical healthcare environments, however, access to data is limited as it Personal Health Information.

\begin{center}
\begin{table}[htbp]
\caption{Comparison with other researches}
    \centering
    \begin{tabular}{llccc}
    \hline
    Researches              & Data                      & Classes & Data-privacy  \\ 
    \hline
    \cite{kim2017diagnosis} & HRV, serum proteomics     & 2     & x    \\
    \cite{byun2019detection}& HRV                       & 2     & x    \\
    \cite{kobayashi2019}    & ECG, HRV, tasks anlaysis  & 2     & x    \\
    \textbf{PFCM } & \textbf{HRV} & \textbf{3} & \textbf{\checkmark}   \\
    
    \hline
    \end{tabular}
    \label{tab1}
\end{table}
\end{center}

\subsection{Federated Learning in Mental Disorder Diagnosis}

Owing to the advantages of FL in privacy protection, there have been studies that have applied FL to monitor and classify mental disorders.
For example, \cite{chhikara2020} used facial expression and speech signals to predict emotion and monitor mental health, and \cite{cao2017deepmood} predicted the severity of the depressive disorder based on users' mobile phone typing dynamics. Reference\cite{xu2021privacy} used both data from users' mobile phone typing and the Hamilton Depression Rating Scale (HDRS) score.
However, these studies did not use HRV data. To the best of our knowledge, no previous studies has used both HRV and FL for MDD analysis.

\section{Personalized Federated Cluster Model}

\subsection{Dataset}

The HRV data used in this study were designed and collected for research at the Department of Psychiatry at Samsung Medical Center. A total of 100 subjects, including a control group, were followed for twelve weeks. Their HRV parameters were measured in three phases: baseline, stress, and recovery. Each subject visited the medical center at least one time and at most five times in the twelve-week time span to provide baseline demographic and clinical data. In addition, non personally identifiable information parameters such as the age and gender of each subject were included in the demographic data. Other scores such as HAM-D, Panic Disorder Severity Scale (PDSS), Anxiety Sensitivity Index (ASI), and Scale for Suicide Ideation (SSI) were also included in the clinical data.

However, mental health scores were excluded from analysis as these metrics directly represent the status of the subject. Our objective is to find potential complex relations between the easily collectable HRV data and depression severity. Thus, our proposed method used only HRV data features and simple biomarker data values as input variables. For severity prediction, we assume HAM-D, in which the scale ranges from 0 to 50, as the target indicator for MDD severity. We categorized the HAM-D values into three classes according to scientific evidence, with 0 to 7 being \textit{Normal}, 8 to 16 being \textit{Mild depression}, and 17 to 50 being \textit{Moderate-Severe depression} \cite{hamilton1960rating}.

Our HRV data involves multi-dimensional features extracted from the time and frequency domains. The standard deviation of average normal to normal intervals (sDNN), number of pairs of adjacent NN intervals differing by more than 50ms in the entire recording (NN50), NN50 count divided by the total number of all NN interval (PNN50), Root Mean Square Differences of successive RR intervals (RMSSD), Very Low Frequency (VLF), Low Frequency (LF), High Frequency (HF), Low/High Frequency ratio (LF/HF), Frequency Power, heart rate, respiration, and temperature variables are collected in our multi-dimensionally extracted HRV features. 

\begin{figure}[h]
\centering
\includegraphics[width=0.95\linewidth]{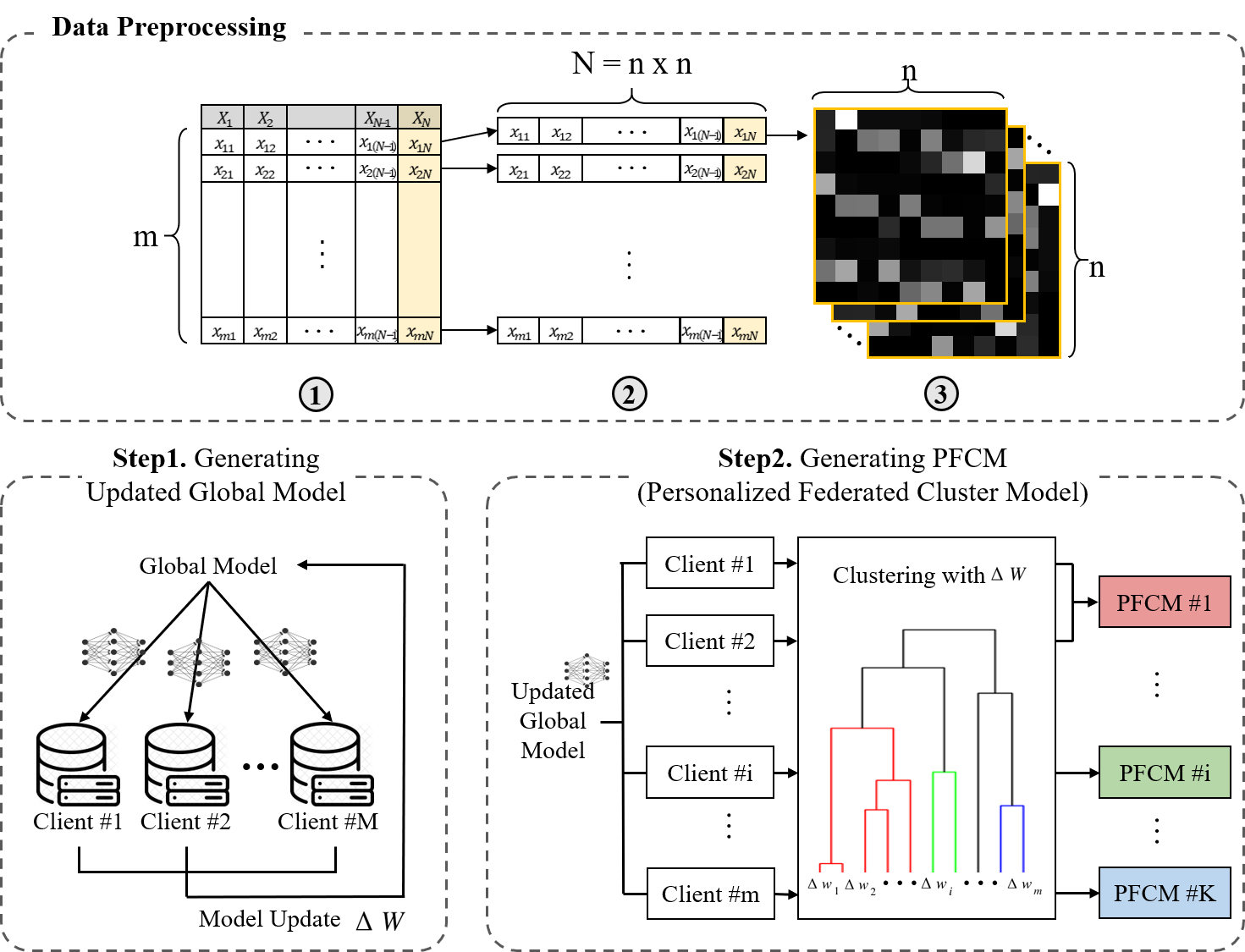}
\caption{An Overview of the Training Procedure of PFCM. 
Top: In data preprocessing step, there are three shapes of data as numbered in this figure. 1) a raw data with null column inserted, 2) a data for each sample, 3) a preprocessed $n \times n$ matrix data for each data sample.
Bottom Left: Step 1 utilizes the conventional FedAvg method for $n$ epochs to generate updated global model. Bottom Right: Step 2 introduces a Personalized Federated Cluster Models (PFCM) created based on the $ \Delta w$ value calculated by comparing each clients with the updated global model.}
\label{fig:architecture}
\end{figure}

\subsection{Preprocessing}

Our HRV dataset has a total of 479 rows and 80 columns, with each row representing a single subject's visit and the columns representing the demographic data and extracted HRV features from three phases (baseline, stress, and recovery). As we use a Convolutional Neural Network (CNN)-based FL framework, we preprocess our data into a $n \times n$ matrix to best exploit the strengths of CNNs to find high-level features. We added an extra dummy column with zero values to create a total of 81 columns. With the 81 columns, we create a $9 \times 9$ matrix based on a single row represented in our HRV dataset as presented at the top of Figure \ref{fig:architecture}. In addition, since the HRV feature variables have different data distributions, we normalized all values between the range of 0 and 1. As such, we created 479 instances of data based on each visit. These 479 instances were then distributed across 100 clients based on the corresponding subject number. To fully simulate a real-world environment, these clients were not given the ability intercommunicate.

\subsection{Training Procedure}

The training procedure is of two steps as shown at the bottom in Figure \ref{fig:architecture}.
In step one, we conduct the conventional FL training method to create an updated global model. 
\begin{equation}\label{eq1}
    W_{t} = \{{w_t}^1, ..., {w_t}^m\}
\end{equation}
The server first distributes a global model across all clients. Received model weights are stored locally as  ${w_t}^1$, ..., ${w_t}^m$ (Eq. \ref{eq1}), where $w_t$ is a global model, and $M$ is the total number of clients.

\begin{equation}\label{eq2}
    \Delta {w_t}^i = {w_t}^i - w_t
\end{equation}
Each client trains its copy of the global model using its local data to calculate the weight difference between the client model and the global model as presented in Eq. \ref{eq2}. Then the updated set of client model weights are saved in $ \Delta W_t $ ($ \Delta W_t = \{ \Delta {w_t}^1, \Delta {w_t}^2, ..., \Delta {w_t}^m\} $).

\begin{equation}\label{eq3-aux}
    \Delta {w_t} = \frac{1}{m}\sum_{i=1}^{m} \Delta {w_t}^i
\end{equation}
After training for $T-1$ times where $T$ is a total number of training epochs prior to clustering, the updated model weights are sent back to the sever. The server aggregates and averages (Eq. \ref{eq3-aux}) the updates from the clients---with the resulting weights symbolized as $ \Delta w_{t+1} $.  

\begin{equation}\label{eq3}
    \Delta w_{t+1} = w_t + \eta_t \Delta {w_t}\\
\end{equation}
The global model is updated based on $ \Delta w_t $, as in Eq. \ref{eq3}. 
Then, the updated global model ($w_T$) is again distributed to each client ($W_{t}$). After this process is repeated for a pre-defined number of rounds, the resulting updated global model is saved for use in step two. 

In step two, we apply hierarchical clustering to the updated global model to create PFCM. Similar to step one, the server first distributes the updated global model to each client. 
Each client trains its copy of the updated global model using its local data. However, rather than aggregating and averaging the weights, the server calculates $ \Delta {w_t}^i $ for each client $i$, defined as the difference between the weights of the updated global model and the weights trained on each client. The resulting $ \Delta {w_t}^i $ for each client $i$ are clustered using the bottom-up hierarchical clustering algorithm to group clients of sufficiently similar weight updates to produce PFCM. Each clustered model is further trained with the conventional FL algorithm described in step one---with a key difference to the number of client updates being aggregated and averaged at each round. The result after a pre-defined number of rounds of training is a PFCM.

\begin{algorithm}
    \SetAlgoLined
    \LinesNotNumbered
    \KwIn{new client with data $D_{test}$}
    Initialize $w_{0}$\\
        \For {$p \gets 0$ \KwTo $P$ }{
            $w_{p+1} \gets WeightUpdate(w_{p})$
        }
        $ \Delta W_{test} \gets w_{P}-w_{T} $ \\
        \For {$k \gets 0$ \KwTo $K$ }{
            $idx \gets FindNearestCluster(C_{k}, \Delta W_{test})$
        }
        Evaluate($W_{C_{idx}}, D_{test}$)
    \caption{Test Procedure. $w_{P}$ is a weight of after $P$ rounds of global model communication on test dataset $D_{test}$, $ \Delta W_{test}$ is a weight difference between the test globally trained model and the global model after post learning, $K$ is a number of clusters, $C_{k}$ is a list of the weights of the cluster models, $idx$ is an index of the nearest cluster, and $W_{C_{idx}}$ is the weight of the nearest cluster model.}\label{alg:test}
\end{algorithm}

\subsection{Testing Procedure}

To fully exploit the strengths of PFCM, we conduct our testing procedure framed as an introduction of new clients, presented in Algorithm \ref{alg:test}. During the testing procedure, we create a queue of clients that did not participate in training. When a new client is introduced, the client is first given the updated global model from step 1 of the training procedure. The client trains this model for a pre-defined number of rounds and sends the weight updates to the server. The server computes $ \Delta W_{test} $, defined as the weights trained on the newly introduced client, and calculates the vector distance between the weights of each cluster model and $ \Delta W_{test} $ using the cosine similarity metric. Then the $idx$ of the most similar cluster model will be found by comparing the cosine similarity between each cluster model's weight and $ \Delta W_{test} $, as represented as $FindNearestCluster(C_{k},\Delta W_{test})$ in Algorithm \ref{alg:test}. Note that we avoid the double-dipping problem; the cluster model used to predict the testing data never sees the testing data. It is a one-time registration process into the PFCM framework wherein the client trains a model using its data and sends the weights of the model to the server. The server then sends the client a cluster model that has not seen the testing data. The client is given the most similar cluster model, with which the accuracy of its data is calculated. After distributing the most similar model to each client in the queue, the accuracy rates are averaged to produce the final test accuracy.

\section{Experiment}

\subsection{Experiment Settings}

\begin{figure}[h]
\centering
\includegraphics[width=\linewidth]{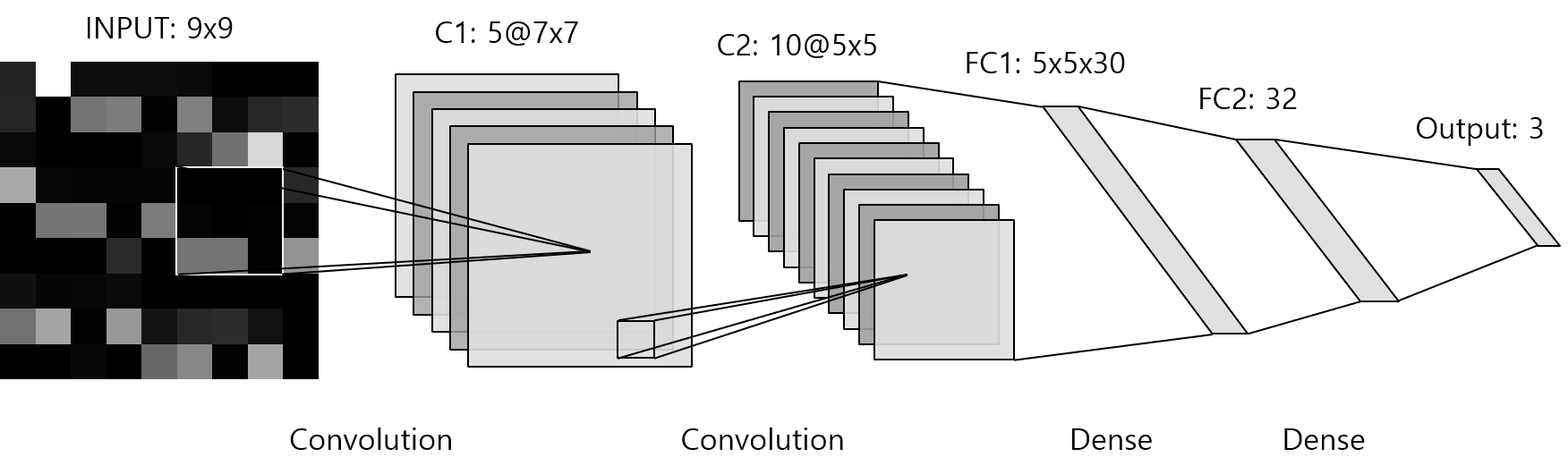}
\caption{Convolutional Neural Network model for preprocessed HRV data.}
\label{fig:hrvcnnmodel}
\end{figure}

Our experiment is framed as a MDD severity classification task with three classes. We used a HRV dataset collected at the Department of Psychiatry at the Samsung Medical Center. There were 479 total samples, each represented by 100 subjects. We created a 80-20 training and testing set split based on the number of subjects.
In addition, we used a Convolutional Neural Network (CNN) model as our base model, shown in Figure \ref{fig:hrvcnnmodel}, and used a self-developed FL framework based on the PyTorch library.

To perform FL on our HRV dataset, we set the input values as $9 \times 9$ matrices. This input was passed through two convolutional layers with 10 and 20 channels, respectively, followed by two fully-connected layers. We trained our initial global model for 50 epochs, using the Stochastic Gradient Descent (SGD) optimizer with a learning rate of 0.1 and a momentum of 0.5, and a the cross-entropy loss function. After creating cluster models using hierarchical clustering, we train each cluster for an additional 20 epochs with the same hyperparameters. The experiment was implemented with Pytorch 1.7.1 and Sklearn 0.24.2, on eight Intel Core i7-10700 processors. GPUs were not required, due to the small size of our dataset.

\begin{figure}[ht]
\centering
\includegraphics[width=0.98\linewidth]{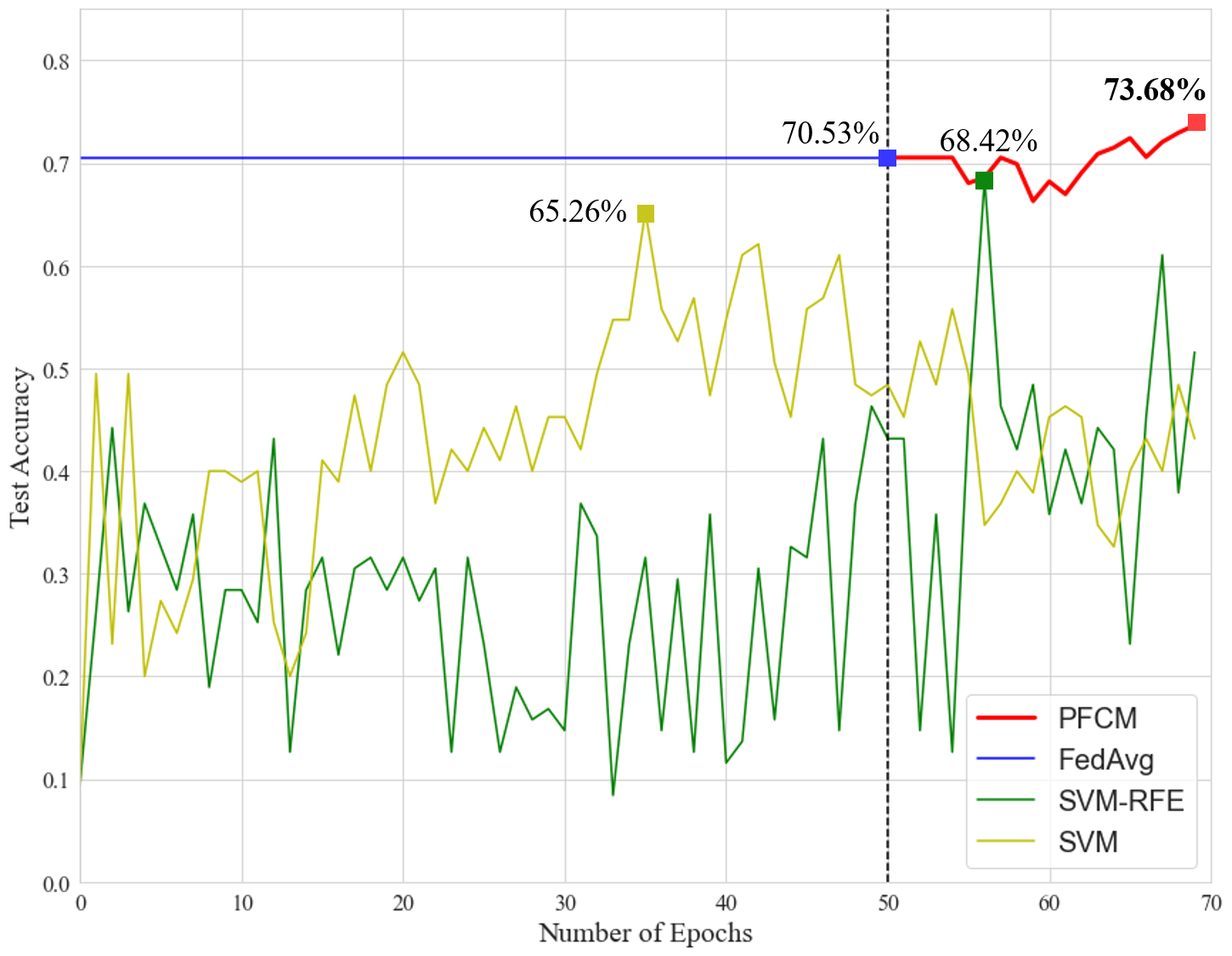}
\caption{A comparison of accuracy between our work (PFCM), FedAvg, and ML frameworks (SVM-RFE and SVM)}
\label{fig:accuracy}
\end{figure}

\subsection{Experiment Results}

In any classification task, it is important to first set a baseline performance by which our methods can be compared. We set our baseline performance to be the popular FedAvg algorithm. In addition, we report results of Support Vector Machine (SVM) and Support Vector Machine with Recursive Feature Elimination (SVM-RFE) method, as it is a method frequently used in related studies. Noted that for SVM and SVM-RFE, the models have access to the entirety of the data, disregarding privacy. Figure \ref{fig:accuracy} shows the accuracy rates on our HRV dataset. For SVM and SVM-RFE, we set the x-axis as the number of maximum iterations. We observed that SVM achieved a lackluster performance of 65.26\%, with SVM-RFE showing an improved performance of 68.42\%. In addition, the model trained on the FedAvg framework showed a stagnant performance of 70.53\%. When our PFCM method was implemented at the end of round 50, marked by the dotted black line, the best performance was seen with an accuracy rate of 73.68\%. Table \ref{table:sens-spec} shows the result of our PFCM in terms of sensitivity and specificity, which are indicators of true positive and true negative rates. Sensitivity is 73.13\% in \textit{Normal} group, 63.16\% in \textit{Mild depression} group, and 100.0\% in \textit{Moderate-Severe depression} group, being more sensitive to classes of higher severity. Specificity is 100.0\% in \textit{Normal} group, 96.05\% in \textit{Mild depression} group, and 74.42\% in \textit{Moderate-Severe depression} group. Overall, our PFCM method achieves better sensitivity and specificity performance compared to studies that analyzed depressive disorder via HRV and other biomarker data. 

\begin{table}[ht]
\caption{Sensitivity and Specificity using PFCM}
    \label{table:sens-spec}

    \centering
    \resizebox{0.9\columnwidth}{!}{%
    \begin{tabular}{lccc} \hline
                & Normal    & Mild        & Moderate-Severe       \\ \hline
    Sensitivity & 73.13\%   & 63.16\%     &  100\%       \\
    Specificity & 100\%     & 96.05\%     &  74.42\%       \\ \hline
    \end{tabular}%
    }
\end{table}

To provide more context regarding the performance of our analysis model, we present our accuracy compared to the results presented in other studies, as shown in Table \ref{table:cmp-accuracy}. Our reported performance of 73.68\% may seem lackluster compared to reported results of other studies, with some reporting accuracy rates of up to 87\%. However, a key distinction is in the number of classes, as the listed studies are based on a 2-class classification task. By changing our task into a 2-class classification task, we also report an increased accuracy rate of 90.70\%. This was the highest accuracy score compared to the results reported in other studies. We want to further the research of MDD analysis by beginning to create more detailed severity classes as opposed to the simple binary classification. Unlike physical illness, most areas of mental disorder, specific diagnosis of severity is more important than determining whether a disease is present.

\begin{center}
\begin{table}[ht]
\caption{Accuracy Comparison with other researches}
    \centering
    \begin{tabular}{lcccc} \hline
    
    Research Paper & Analysis Method & Number of Classes & Accuracy         \\ \hline
    \cite{byun2019detection}& SVM-RFE       & 2     & 74.40\%               \\
    \cite{kim2017diagnosis} & SVM-RFE       & 2     & 80.10\%               \\
    \cite{kobayashi2019}    & SVM+PCA       & 2     & 87.00\%               \\
    \textbf{PFCM} & \textbf{CNN-based FL} & \textbf{2} & \textbf{90.70\%}   \\
    PFCM          & CNN-based FL  & 3   & 73.68\%                           \\ \hline
    \end{tabular}
    \label{table:cmp-accuracy}
\end{table}
\end{center}

\section{Discussion}

Artificial Intelligence is continually spreading to many specific fields of healthcare with the increase of data quantity and quality, paired with developing computing power. There have been various attempts to find implicit and complex relations between variables that were once seen to have limited correlation. With MDD in particular, promising researches have been published regarding the increase in analysis performance using ML. Practical limitations of these methods, however, are data privacy laws that restrict the collection of healthcare data. Certain researches that have addressed the issue of privacy failed to show adequate results in realistic non-IID scenarios. 

Our research have shown that it is possible to successfully apply ML algorithms on privately distributed HRV data. PFCM shows a result of 78.76\% and 90.16\% for average sensitivity and specificity, respectively. This is an increased performance compared to the baseline method of FedAvg. This was a result of an improved performance when personalized models were distributed to each client. A root cause of the stagnant performance seen with FL is highly skewed data. Despite extended training, the model continued to predict that every input was of a normal subject. In a non-IID scenario, one in which the labels of the data of each client are heavily skewed, vanilla FedAvg algorithms creates a highly biased model. Given this type of imbalance, a common solution would be to re-sample or give higher weights to classes with less samples. However, this method is impossible to implement without access to data. By creating models with different biases solely based on local weight updates, and by implementing a clustering phase at the introduction of a new client, we can provide each client with a correctly biased model---a model that is biased towards the corresponding client's data. 

In addition, PFCM showed higher accuracy compared to SVM and SVM-RFE, despite holding the assumption of accessible and centralized data. Though SVMs are a powerful tool, lower performance was observed as it was unable to make clear distinctions between \textit{Normal} and \textit{Mild} subjects. Owing to its ability to extract high-level features, the CNN-based approach made clearer distinctions between the severity classes, achieving higher performance. This is particularly important, as PFCM showed better performance while conforming to privacy regulations.

The significance of our PFCM method is apparent when compared to the results reported in other studies. Though other related studies report higher accuracy rates on a 2-class classification task, these are unfair comparisons to our research, as we report our accuracy rates on a 3-class classification task. However, if we change our dataset into a 2-class classification task, by combining the \textit{Normal} and \textit{Mild} classes, we report accuracy rates that are higher than previously reported studies. Not only does this demonstrate that our method shows superior performance, it also shows that the HRV of \textit{Normal} subjects are almost indistinguishable with \textit{Mild} subjects.

\subsection*{Limitations \& Future Works}

The main limitation to this study is the lack of data. Though it would have been optimal to separate our data into four classes based on the HAM-D metric, due to unavailability of data, we combined the categories of \textit{Moderate} and \textit{Severe}. Furthermore, we have trained our model to be more sensitive to higher levels of HAM-D. This resulted in a decreased sensitivity of normal subjects. For this framework to be adopted as a screening test in a practical healthcare setting, higher sensitivity of all classes are required. However, we believe these limitations can be solved with more data, especially for \textit{Moderate} and \textit{Severe} patients.

In future studies, we plan to conduct further clinical trials with Samsung Medical Center. This will allow collection and analysis of data containing more variables. In addition to data collection, we also plan to improve mental disorder diagnostic performance by exploring additional applications of ML techniques such as Recurrent Neural Network, Long-Short Term Memory, and Generative Adversarial Network in our analysis. 

\section{Conclusion}
We have shown that even without a large and balanced HRV dataset, our Personalized Federated Cluster Model (PFCM) can accurately analyze the severity of a Major Depressive Disorder (MDD). In addition, we have shown that current Federated Learning methods can outperform other machine learning techniques through proper clustering of sufficiently similar clients. Lastly, unlike previous researches, our PFCM method provides a solution to apply ML techniques in strict data privacy environments. With a higher quantity and balance of data, we believe it may be viable to use PFCM to enable analysis of not only many problems in the field of healthcare, but also in many realistic FL environments.

\section*{Acknowledgement}
This work was partly supported by Institute of Information \& communications Technology Planning \& Evaluation (IITP) grant funded by the Korea government(MSIT) (No.2019-0-00421, Artificial Intelligence Graduate School Program(Sungkyunkwan University)), and Healthcare AI Convergence Research \& Development Program through the National IT Industry Promotion Agency of Korea (NIPA) funded by Ministry of Science and ICT (No. S1601-20-1041)

\bibliographystyle{IEEEtran}
\bibliography{bibtemplate}

\begin{thebibliography}{10}
\providecommand{\url}[1]{#1}
\csname url@samestyle\endcsname
\providecommand{\newblock}{\relax}
\providecommand{\bibinfo}[2]{#2}
\providecommand{\BIBentrySTDinterwordspacing}{\spaceskip=0pt\relax}
\providecommand{\BIBentryALTinterwordstretchfactor}{4}
\providecommand{\BIBentryALTinterwordspacing}{\spaceskip=\fontdimen2\font plus
\BIBentryALTinterwordstretchfactor\fontdimen3\font minus
  \fontdimen4\font\relax}
\providecommand{\BIBforeignlanguage}[2]{{%
\expandafter\ifx\csname l@#1\endcsname\relax
\typeout{** WARNING: IEEEtran.bst: No hyphenation pattern has been}%
\typeout{** loaded for the language `#1'. Using the pattern for}%
\typeout{** the default language instead.}%
\else
\language=\csname l@#1\endcsname
\fi
#2}}
\providecommand{\BIBdecl}{\relax}
\BIBdecl

\bibitem{chen2020deep}
Y.-W. Chen and L.~C. Jain, ``Deep learning in healthcare,'' 2020.

\bibitem{tobore2019health}
\BIBentryALTinterwordspacing
I.~Tobore, J.~Li, L.~Yuhang, Y.~Al-Handarish, A.~Kandwal, Z.~Nie, and L.~Wang,
  ``Deep learning intervention for health care challenges: Some biomedical
  domain considerations,'' \emph{JMIR Mhealth Uhealth}, vol.~7, no.~8, p.
  e11966, Aug 2019. [Online]. Available:
  \url{http://www.ncbi.nlm.nih.gov/pubmed/31376272}
\BIBentrySTDinterwordspacing

\bibitem{houser2018gdpr}
K.~A. Houser and W.~G. Voss, ``Gdpr: The end of google and facebook or a new
  paradigm in data privacy,'' \emph{Rich. JL \& Tech.}, vol.~25, p.~1, 2018.

\bibitem{belmaker2008major}
R.~Belmaker and G.~Agam, ``Major depressive disorder,'' \emph{New England
  Journal of Medicine}, vol. 358, no.~1, pp. 55--68, 2008.

\bibitem{mcmahan2017communication}
B.~McMahan, E.~Moore, D.~Ramage, S.~Hampson, and B.~A. y~Arcas,
  ``Communication-efficient learning of deep networks from decentralized
  data,'' in \emph{Artificial Intelligence and Statistics}.\hskip 1em plus
  0.5em minus 0.4em\relax PMLR, 2017, pp. 1273--1282.

\bibitem{konevcny2016federated}
J.~Kone{\v{c}}n{\`y}, H.~B. McMahan, F.~X. Yu, P.~Richt{\'a}rik, A.~T. Suresh,
  and D.~Bacon, ``Federated learning: Strategies for improving communication
  efficiency,'' \emph{arXiv preprint arXiv:1610.05492}, 2016.

\bibitem{Li_2020}
\BIBentryALTinterwordspacing
T.~Li, A.~K. Sahu, A.~Talwalkar, and V.~Smith, ``Federated learning:
  Challenges, methods, and future directions,'' \emph{IEEE Signal Processing
  Magazine}, vol.~37, no.~3, p. 50–60, May 2020. [Online]. Available:
  \url{http://dx.doi.org/10.1109/MSP.2020.2975749}
\BIBentrySTDinterwordspacing

\bibitem{sattler2020clustered}
F.~Sattler, K.-R. M{\"u}ller, and W.~Samek, ``Clustered federated learning:
  Model-agnostic distributed multitask optimization under privacy
  constraints,'' \emph{IEEE Transactions on Neural Networks and Learning
  Systems}, 2020.

\bibitem{briggs2020federated}
C.~Briggs, Z.~Fan, and P.~Andras, ``Federated learning with hierarchical
  clustering of local updates to improve training on non-iid data,'' 2020.

\bibitem{kim2017diagnosis}
E.~Y. Kim, M.~Y. Lee, S.~H. Kim, K.~Ha, K.~P. Kim, and Y.~M. Ahn,
  ``{{D}iagnosis of major depressive disorder by combining multimodal
  information from heart rate dynamics and serum proteomics using
  machine-learning algorithm},'' \emph{Prog Neuropsychopharmacol Biol
  Psychiatry}, vol.~76, pp. 65--71, 06 2017.

\bibitem{byun2019detection}
S.~Byun, A.~Y. Kim, E.~H. Jang, S.~Kim, K.~W. Choi, H.~Y. Yu, and H.~J. Jeon,
  ``Detection of major depressive disorder from linear and nonlinear heart rate
  variability features during mental task protocol,'' \emph{Computers in
  biology and medicine}, vol. 112, p. 103381, 2019.

\bibitem{kobayashi2019}
M.~{Kobayashi}, G.~{Sun}, T.~{Shinba}, T.~{Matsui}, and T.~{Kirimoto},
  ``Development of a mental disorder screening system using support vector
  machine for classification of heart rate variability measured from
  single-lead electrocardiography,'' in \emph{2019 IEEE Sensors Applications
  Symposium (SAS)}, 2019, pp. 1--6.

\bibitem{chhikara2020}
P.~{Chhikara}, P.~{Singh}, R.~{Tekchandani}, N.~{Kumar}, and M.~{Guizani},
  ``Federated learning meets human emotions: A decentralized framework for
  human-computer interaction for iot applications,'' \emph{IEEE Internet of
  Things Journal}, pp. 1--1, 2020.

\bibitem{cao2017deepmood}
\BIBentryALTinterwordspacing
B.~Cao, L.~Zheng, C.~Zhang, P.~S. Yu, A.~Piscitello, J.~Zulueta, O.~Ajilore,
  K.~Ryan, and A.~D. Leow, ``Deepmood: Modeling mobile phone typing dynamics
  for mood detection,'' in \emph{Proceedings of the 23rd ACM SIGKDD
  International Conference on Knowledge Discovery and Data Mining}, ser. KDD
  '17.\hskip 1em plus 0.5em minus 0.4em\relax New York, NY, USA: Association
  for Computing Machinery, 2017, p. 747–755. [Online]. Available:
  \url{https://doi.org/10.1145/3097983.3098086}
\BIBentrySTDinterwordspacing

\bibitem{xu2021privacy}
X.~Xu, H.~Peng, L.~Sun, M.~Z.~A. Bhuiyan, L.~Liu, and L.~He,
  ``Privacy-preserving federated depression detection from multi-source mobile
  health data,'' 2021.

\bibitem{hamilton1960rating}
M.~Hamilton, ``A rating scale for depression,'' \emph{Journal of neurology,
  neurosurgery, and psychiatry}, vol.~23, no.~1, p.~56, 1960.

\end{thebibliography}

\end{document}